\pdfoutput=1
\documentclass[conference]{IEEEtran}
\IEEEoverridecommandlockouts
\usepackage[T1]{fontenc}
\usepackage{cite}
\usepackage{booktabs}
\usepackage{multirow}
\usepackage{amsmath,amssymb,amsfonts,mathtools,bm}
\usepackage{graphicx}
\usepackage{subfig}
\usepackage[font=small]{caption}
\usepackage{algorithm}
\usepackage{algpseudocode}
\usepackage{xcolor}
\usepackage[hidelinks]{hyperref}

\usepackage{mathrsfs}
\usepackage{cleveref}

\newcommand{\fameqp}{\ensuremath{\mathrm{FAME}\text{-}\mathrm{QP}}}
\newcommand{\mbb}[1]{\mathbb{#1}}
\newcommand{\R}{\mathbb{R}}
\newcommand{\N}{\mathbb{N}}
\def\bs{\boldsymbol}
\def\mf{\mathbf}
\def\mc{\mathcal}
\def\mcr{\mathscr}
\DeclareMathOperator*{\argmin}{arg\,min}
\DeclareMathOperator{\diag}{diag}
\newcommand{\proj}{\Pi}
\newcommand{\alnn}[1]{\begin{align}#1\end{align}}
\newcommand{\salnn}[2]{\begin{subequations}\label{#1}\begin{align}#2\end{align}\end{subequations}}

\makeatletter
\let\OldStatex\Statex
\renewcommand{\Statex}[1][3]{%
  \setlength\@tempdima{\algorithmicindent}%
  \OldStatex\hskip\dimexpr#1\@tempdima\relax}
\makeatother

\providecommand{\SI}[2]{\ensuremath{#1\,\mathrm{#2}}}
\providecommand{\SIrange}[3]{\ensuremath{#1\text{--}#2\,\mathrm{#3}}}
\providecommand{\second}{s}
\providecommand{\milli}{m}
\providecommand{\giga}{G}
\providecommand{\byte}{B}
\providecommand{\mega}{M}
\providecommand{\hertz}{Hz}
\providecommand{\meter}{m}
\providecommand{\radian}{rad}

\allowdisplaybreaks
\title{\LARGE \fameqp: A GPU-Accelerated
Quadratic-Programming Solver for Real-Time Model Predictive Control}

\author{Santosh Mohan Rajkumar\thanks{
Authors are with the Department of Mechanical \& Aerospace Engineering, The Ohio State University, Columbus, OH, USA. E-mail:\{\texttt{rajkumar.36, goswami.78, stockar.1}\}\texttt{@osu.edu}
},
Debdipta Goswami, and 
Stephanie Stockar\thanks{
Open Source Code: \url{https://github.com/santoshrajkumar/fame-qp}
}
}

\begin{document}
\bstctlcite{IEEEexample:BSTcontrol}

\maketitle
\thispagestyle{empty}
\pagestyle{empty}

\begin{abstract}
We present FAME-QP, a GPU-accelerated solver for dense quadratic programs arising in linear model predictive control. FAME-QP combines constraint-dependent operator representations with fused single- and two-kernel iterations replayed through CUDA graphs, automatically selecting the execution configuration through device calibration and problem-specific timing. It probes a cached operator bank to select the initial penalty and updates operators lazily when problem matrices change. Numerical benchmarks and simulated humanoid and quadrotor control experiments demonstrate reduced latency relative to competing GPU solvers while maintaining solution accuracy and control performance.
\end{abstract}

\section{Introduction}

Quadratic programming (QP) problems arise in a wide range of applications, including statistics, machine learning, finance, and model predictive control (MPC) for robotic and control systems \cite{turlach2015quadratic, stellato2020osqp, wensing2023optimization}. They also appear as subproblems in nonlinear optimization algorithms, such as sequential quadratic programming (SQP). Many of these applications lead to convex QP problems. Among these applications, MPC requires solving an optimization problem within each control sampling period, making computational efficiency critical for real-time operation \cite{wang2010fast}. With linear prediction models, convex quadratic costs, and linear constraints, each MPC update requires solving a convex QP problem, yielding a formulation commonly referred to as linear MPC (LMPC). 

Common solution approaches for QP problems include interior-point (IP), active-set (AS), operator-splitting (OS), and proximal methods.
Representative IP methods include HPIPM \cite{frison2020hpipm}, qpSWIFT \cite{pandala2019qpswift}, PIQP \cite{schwan2023piqp}, and Clarabel \cite{goulart2026clarabel} (with QP as a special case). Although IP methods often converge in relatively few iterations, each iteration requires solving a potentially expensive linear system. They are also generally more difficult to warm start, limiting their ability to exploit similarities between successive QP problems in MPC \cite{bishop2024reluqp}. 

AS methods, such as qpOASES \cite{ferreau2014qpoases} and DAQP \cite{arnstrom2022daqp}, readily support warm starting and can be highly efficient when initialized with the correct active set. However, their combinatorial worst-case complexity in the number of constraints can limit their scalability to large-scale QPs \cite{bishop2024reluqp}. OS methods based on the alternating direction method of multipliers (ADMM), such as OSQP \cite{stellato2020osqp}, and related proximal methods, such as ProxQP \cite{bambade2022proxqp}, are increasingly used in robotic MPC owing to their ease of warm starting and suitability for applications requiring moderate solution accuracy \cite{schubiger2020cuosqp,dinh2025real}. Moreover, their computational structure offers opportunities for parallelization. Nevertheless, their convergence rates can be slower than those of other methods, particularly when high-accuracy solutions are required \cite{schwan2023piqp}.
\begin{figure}[htpb!]
    \centering
    \includegraphics[width=0.99\columnwidth]{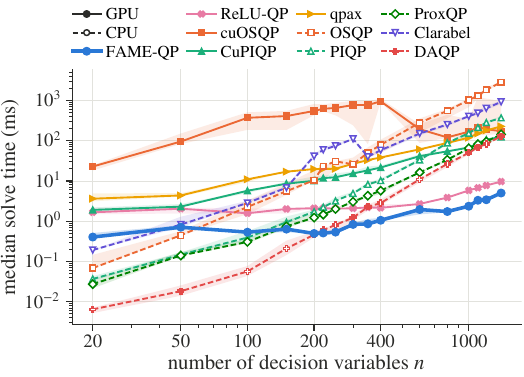}
    \caption{Cold-solve times on dense random QPs at matched accuracy,
excluding setup. Lines show medians and bands min--max ranges across
five instances, each timed by its median over repeated cold solves.
All solvers use \texttt{float64} except cuOSQP (\texttt{float32}).}
    \label{fig:random_qp}
    \vspace{-0.8em}
\end{figure}

Although most established QP solvers are CPU-based, GPUs offer opportunities to accelerate large-scale QP solving through massive parallelism and higher memory bandwidth, enabling concurrent execution of independent computations and faster data movement \cite{schubiger2020cuosqp}. Growing interest in MPC for robotics and control using world models \cite{hansen2024td, rajkumar2026kpwm}, locally linear latent dynamics \cite{watter2015embed}, and Koopman-based linear, bilinear, or semi-linear representations \cite{korda2018linear,goswami2021bilinearization,rajkumar2025real} further motivates GPU acceleration to meet the real-time computational demands of large-scale MPCs. Moreover, modern robotics increasingly relies on GPUs for parallel simulation, inference, and learning \cite{makoviychuk2021isaac,rudin2022learning,oneill2024open}. Integrating CPU-native QP solvers into such pipelines can introduce additional latency through frequent CPU–GPU data transfers \cite{lu2025mpax}. Together, these considerations motivate GPU-native QP solvers that keep optimization data and computations on the device.

This work presents \fameqp{}, \textbf{F}used \textbf{A}DMM with \textbf{M}ultimode \textbf{E}xecution for \textbf{Q}P solution, a GPU solver for the dense QPs of linear MPC. Its contributions are: (i) constraint-dependent operator representations that reduce operator storage by up to $4\times$ and iteration time by up to $3\times$, relative to treating all constraint rows as dense; (ii) fused single- and two-kernel iterations with CUDA graph replay and automatic selection using device calibration and problem-specific timing; (iii) initial penalty selection through cached-operator probing and lazy operator updates for changing QP matrices; (iv) numerical and simulated robotic MPC evaluations demonstrating lower latency than the competing GPU methods evaluated in each benchmark, with matched solution accuracy on random QPs and comparable closed-loop control performance; and (v) an open-source implementation.

\subsection{Related Work}

\noindent\textbf{GPU QP solvers.}
cuOSQP~\cite{schubiger2020cuosqp} parallelizes OSQP through indirect linear solves, while qpax~\cite{tracy2024qpax} and CuPIQP~\cite{song2026cupiqp} implement IP methods on GPUs, with CuPIQP exploiting kernel fusion and CUDA graph replay.
The closest approach is ReLU-QP~\cite{bishop2024reluqp}, which reformulates ADMM as a weight-tied neural network and caches pre-inverted operators for penalty adaptation.
Its dense operator and repeated dispatch introduce computational and execution overhead, while its offline precomputation limits online matrix updates.
\fameqp{} retains the pre-inversion and penalty-bank approach but uses smaller, constraint-dependent operators, fused graph-replayed iterations, and lazy operator updates.

\noindent\textbf{GPU MPC solvers.}
MPCGPU~\cite{adabag2024mpcgpu} and TurboMPC~\cite{bravo2026turbompc} accelerate nonlinear MPC through GPU implementations of SQP and structured linear algebra.
$\pi$MPC~\cite{wu2026pimpc} exploits parallelism across the prediction horizon through consensus ADMM, while CUDA MPC~\cite{akbari2026cudampc} reduces execution overhead using a persistent cooperative kernel and CUDA graphs.
These methods exploit structure at the optimal-control level, whereas \fameqp{} accepts dense convex QPs directly, including general constraint matrices and online matrix changes.

\noindent\textbf{Positioning.}
Our contribution lies in combining constraint-dependent operator representations with automatic selection between fused single- and two-kernel iterations, initial penalty probing, and lazy matrix updates.
The execution configuration is selected through device calibration and problem-specific timing, targeting the latency of individual MPC solves.
\section{Preliminaries}
\label{sec:background}

\noindent \textbf{Notations.} Define $\N_0 := \{0\} \cup \N$. Let $\mbb S^n_+$ be the set of symmetric positive semi-definite matrices of dimension $n\times n$.  For $a,b \! \in \! \N_0$ with $b\!\ge\! a$, define $[a:b]:=\{a,\ldots,b \}$. Let $\mf 0$ and $I$ denote the zero vector or matrix and the identity matrix, respectively, of appropriate dimensions. For a matrix or vector, a space or comma separates columns, and a semicolon separates rows, where applicable.

\noindent \textbf{Convex QP problem.} A convex QP problem takes the following form \cite{borrelli2017predictive}
\alnn{
\min_{z \in \R^n} \tfrac12 z^\top P z + q^\top z
  \quad\text{s.t.}\quad l \le G z \le h,
  \label{eq:qp}
}
where $z $ is the decision variable, $P \in \mbb S^n_+$, $G \in \R^{m\times n}$, $q \in \R^n$, $l\in(\R \cup \{-\infty \})^m$, and  $h\in(\R \cup \{+\infty \})^m$.

\noindent \textbf{LMPC.}
Consider the discrete-time linear system $x_{k+1}=Ax_k+Bu_k$,
where $x_k\! \in \!\R^{n_x}$ and $u_k\! \in \! \R^{n_u}$ denote the state
and input, respectively. At each sampling instant, LMPC measures
the current state $\hat x$ and solves the following optimal control problem (OCP) over a prediction horizon of $N$ steps
\cite{borrelli2017predictive}:
\salnn{eq:lmpc}{
  \min_{x,u}\;& \tfrac12\textstyle\sum_{k=0}^{N-1}
      \bigl(x_k^\top Q x_k + u_k^\top R u_k\bigr)
      + \tfrac12 x_N^\top Q_N x_N\\
  \text{s.t.}\;& x_0=\hat x,\ x_{k+1}=Ax_k+Bu_k,
      \ k \in[0:N-1],\\
  & \underline u\le u_k\le\overline u,
      \quad k \in[0:N-1],\\
  & \underline x\le x_k\le\overline x,
      \quad k \in[1:N],
}
with $x=\{ x_k \}_{k=0}^N$ and $u=\{u_k\}_{k=0}^{N-1}$. Here, $Q,Q_N\succeq0$ and $R\succ0$ are the state, terminal,
and input weighting matrices, respectively. The controller applies
the first optimal input $u_0^\star$, shifts the horizon forward
by one step, and repeats the procedure at the next sampling instant.

\noindent\textbf{GPU computation.} We restrict attention to NVIDIA GPUs and the CUDA programming model \cite{nvidia2025cudaguide}. GPU computations execute as \emph{kernels}: functions typically launched by the host and executed in parallel by threads grouped into blocks \cite{nickolls2008scalable}. Threads within a block share fast on-chip memory and synchronize efficiently, while communication between blocks typically uses global memory. Successive kernels in the same stream execute in order, making kernel boundaries a standard way to synchronize all blocks between computational stages \cite{nvidia2025cudaguide}. A \emph{CUDA graph} represents a workflow of GPU operations, such as kernel
launches and data transfers, and the dependencies between them.
The workflow is prepared once and can then be executed repeatedly with a single host call, reducing host-side launch overhead \cite{nvidia2025cudaguide}.

When a kernel is launched, its thread blocks are scheduled onto
the GPU's streaming multiprocessors (SMs), where threads execute
in groups of 32 called warps, using values held in registers.
Global memory accesses may be served by GPU caches, while
threads within a block can exchange data through on-chip shared
memory. Registers and shared memory have thread and block
lifetimes, respectively, so intermediate results passed between
kernels are typically stored in global memory
\cite{nvidia2025cudaguide}. The runtime of a computation expressed
as a sequence of kernels therefore depends on arithmetic, data
movement through the memory hierarchy, and launch and scheduling
overhead. \emph{Kernel fusion} combines operations that would
otherwise execute in separate kernels into a single kernel,
allowing some intermediate results to remain in registers or
shared memory. This can reduce both the number of kernel
launches and global memory accesses \cite{filipovic2015fusion}.

\subsection{Solving the QP with ADMM}

To solve \eqref{eq:qp}, we employ the ADMM formulation proposed in \cite{stellato2020osqp}. Introducing an auxiliary
variable $s\in\R^m$ yields the equivalent problem
\alnn{
\min_{z,s} \tfrac12 z^\top P z + q^\top z + \mcr I_{\mc C}(s)
  \quad\text{s.t.}\quad Gz = s,
  \label{eq:qp1}
}
where $\mcr I_{\mc C}(s) = \{ 0 \text{ if } s \in \mc C,\ +\infty \text{ otherwise} \}$ and $\mc C:=\{ s:\ l\le s\le h \}$. 
With auxiliary variables $\tilde z \in \R^n$ and $\tilde s \in \R^m$, \eqref{eq:qp1} can be rewritten as
\salnn{eq:qp2}{
\min_{z,s, \tilde s, \tilde z}\  & \mc J =  \tfrac12 \tilde z^\top P \tilde z + q^\top \tilde z + \mcr I_{\mc C}(s)+ \mcr I_{\mc E}(\tilde z, \tilde s)  \\
  &\text{s.t.}\quad (\tilde z, \tilde s) = (z,s),
}
where $\mcr I_{\mc E}(\tilde z, \tilde s) = \{ 0 \text{ if } G\tilde z = \tilde s,\ +\infty \text{ otherwise} \}$ and $\mathcal E:=\{(\tilde z,\tilde s):\tilde s=G\tilde z\}$. Let $\mu\in\R^n$ and $\lambda\in\R^m$ be the Lagrange
multipliers associated with $\tilde z=z$ and $\tilde s=s$,
respectively. The augmented Lagrangian, omitting terms constant in the primal variables, is
\alnn{
\mc L = \mc J + \tfrac\sigma 2\| \tilde z - z+\mu/\sigma\|_2^2 + \tfrac12 \| \tilde s -s + \bs \rho^{-1}\lambda \|^2_{\bs \rho}
}
where $\sigma>0$ is the proximal regularization parameter, $\bs \rho=\operatorname{diag}(\rho_1,\ldots,\rho_m)
\succ0$ is the diagonal penalty matrix, and $\|v\|_{\bs \rho}^2=v^\top \bs \rho v$.

Given the iterate $(z^j,s^j,\mu^j,\lambda^j)$ at iteration $j\in\N_0$ and a relaxation parameter $\alpha\in(0,2)$, ADMM involves the following steps
\salnn{eq:admm}{
(\tilde z^{j+1},\tilde s^{j+1})
&= \argmin_{\tilde z,\tilde s}\,
\mc L(\tilde z,\tilde s,z^j,s^j;\mu^j,\lambda^j),
\label{eq:admm_eq}\\
z^{j+1}
&= \alpha\tilde z^{j+1}+(1-\alpha)z^j
+\sigma^{-1}\mu^j,
\label{eq:admm_z}\\
s^{j+1}
&= \argmin_{s\in\mc C}\,
\tfrac12\|\hat s^{j+1}-s
+\boldsymbol{\rho}^{-1}\lambda^j
\|_{\boldsymbol{\rho}}^2,
\label{eq:admm_s}\\
\mu^{j+1}
&= \mu^j+\sigma\bigl(
\alpha\tilde z^{j+1}+(1-\alpha)z^j-z^{j+1}\bigr),
\label{eq:admm_mu}\\
\lambda^{j+1}
&= \lambda^j+\boldsymbol{\rho}
\bigl(\hat s^{j+1}-s^{j+1}\bigr),
\label{eq:admm_lambda}
}
where $\hat s^{j+1}:=\alpha\tilde s^{j+1}+(1-\alpha)s^j$.
The $z$-update implies $\mu^{j+1}=\mf 0$; hence,
initializing $\mu^0=\mf 0$ eliminates $\mu$ from all iterations.

\noindent\textbf{Closed form.}
Substituting $\tilde s=G\tilde z$ into
\eqref{eq:admm_eq} and setting the gradient with respect to
$\tilde z$ to zero yields
\alnn{
K\tilde z^{j+1}
=\sigma z^j-\mu^j-q
+G^\top(\boldsymbol{\rho}s^j-\lambda^j),
}
where $K:=P+\sigma I+G^\top\boldsymbol{\rho}G\succ0$.
Moreover, since
$\boldsymbol{\rho}$ is positive diagonal and $\mc C$ is a box,
\eqref{eq:admm_s} separates into scalar minimizations solved
by componentwise clipping. With $\mu^0=\mf 0$ and $s^0\in\mc C$, one ADMM iteration reduces to
\salnn{eq:admm2}{
\tilde z^{j+1}
&= M [\sigma z^j-q
+G^\top(\boldsymbol{\rho}s^j-\lambda^j)],
\label{eq:linsys}\\
\tilde s^{j+1}&=G\tilde z^{j+1},\label{eq:linsys2} \\
z^{j+1}
&=\alpha\tilde z^{j+1}+(1-\alpha)z^j, \\
\zeta^{j+1}
&=\alpha\tilde s^{j+1}+(1-\alpha)s^j
+\boldsymbol{\rho}^{-1}\lambda^j, \\
s^{j+1}&=\proj_{\mc C}(\zeta^{j+1}), \\
\lambda^{j+1}
&=\boldsymbol{\rho}(\zeta^{j+1}-s^{j+1}),
}
where $\proj_{\mc C}(\zeta)=\min\{\max\{\zeta,l\},h\}$
acts componentwise and $M=K^{-1}$. Here, $\zeta^{j+1}$ is the unconstrained
minimizer of the quadratic in \eqref{eq:admm_s}, and
$\boldsymbol{\rho}^{-1}\lambda^{j+1}$ is the projection
residual. Consequently, $\lambda_i^{j+1}>0$ implies
$s_i^{j+1}=h_i$, $\lambda_i^{j+1}<0$ implies
$s_i^{j+1}=l_i$, and $l_i<s_i^{j+1}<h_i$ implies
$\lambda_i^{j+1}=0$.

\noindent\textbf{Optimality and termination.}
At a fixed point of \eqref{eq:admm2},
$\tilde z=z$ and $\tilde s=s=Gz$, yielding
$Pz+q+G^\top\lambda=\mf 0$ and
$s=\proj_{\mc C}(s+\boldsymbol{\rho}^{-1}\lambda)$.
These are the KKT conditions of \eqref{eq:qp};
$\sigma$ and $\boldsymbol{\rho}$ affect convergence
but not these conditions. The iteration terminates when
\salnn{eq:stop}{
r_{\rm prim}^j:=\|Gz^j-s^j\|_\infty
&\le \epsilon_{\rm abs}+\epsilon_{\rm rel}\eta_{\rm p}^j,\\
r_{\rm dual}^j:=\|Pz^j+q+G^\top\lambda^j\|_\infty
&\le \epsilon_{\rm abs}+\epsilon_{\rm rel}\eta_{\rm d}^j,
}
where
$\eta_{\rm p}^j:=\max\{\|Gz^j\|_\infty,\|s^j\|_\infty\}$,
$\eta_{\rm d}^j:=\max\{\|Pz^j\|_\infty,
\|G^\top\lambda^j\|_\infty,\|q\|_\infty\}$,
and $\epsilon_{\rm abs},\epsilon_{\rm rel}>0$
are the absolute and relative tolerances, respectively.

\subsection{LMPC as a QP Problem}

With the predicted states and inputs stacked into the decision vector $z=\operatorname{col}(x_0,u_0,\ldots,x_{N-1},u_{N-1},x_N)$, the OCP~\eqref{eq:lmpc} admits a sparse QP formulation of the general form~\eqref{eq:qp} \cite[Sec. 14.2.2]{borrelli2017predictive}.
The sparse QP formulation of LMPC has $n=(N+1)n_x+Nn_u$ decision variables and $m=Nn_x+n$ constraint rows. Eliminating the predicted states through the dynamics yields a condensed QP formulation of~\eqref{eq:lmpc} with only $n=Nn_u$ input variables \cite{jerez2011condensed}, whose dense matrix structure is well suited to GPU-accelerated operations in standard machine-learning libraries. Let $U=\operatorname{col}(u_0,\ldots,u_{N-1})$ and
$X=\operatorname{col}(x_1,\ldots,x_N)$. The linear dynamics give
$X=\mathcal A\hat x+\mathcal B U$, where
\alnn{
\mathcal A=
\begin{bmatrix}
\scriptstyle A\\
\scriptstyle A^2\\
\scriptstyle \vdots\\
\scriptstyle A^N
\end{bmatrix} ,
\qquad
\mathcal B=
\begin{bmatrix}
\scriptstyle B & \scriptstyle \mf 0 & \scriptstyle \cdots & \scriptstyle \mf 0\\
\scriptstyle AB & \scriptstyle B & \scriptstyle \cdots & \scriptstyle \mf 0\\
\scriptstyle \vdots & \scriptstyle \vdots & \scriptstyle \ddots & \scriptstyle \vdots\\
\scriptstyle A^{N-1}B & \scriptstyle A^{N-2}B & \scriptstyle \cdots & \scriptstyle B
\end{bmatrix},
}
with $\mathcal A\in\R^{Nn_x\times n_x}$ and $\mathcal B\in\R^{Nn_x\times Nn_u}$.
With $\mathcal Q\!=\!\operatorname{blkdiag}(Q,\ldots,Q,Q_N)\in\R^{Nn_x\times Nn_x}$
and $\mathcal R\!=\!\operatorname{blkdiag}(R,\ldots,R)$,
the condensed QP takes the form~\eqref{eq:qp} with $z=U$ and
\alnn{
& P=\mathcal B^\top\mathcal Q\mathcal B+\mathcal R,
\quad q=\mathcal B^\top\mathcal Q\mathcal A\hat x,
\quad G=[\mathcal B;\ I], \notag\\
& l=[\underline X-\mathcal A\hat x;\ \underline U],
\quad h=[\overline X-\mathcal A\hat x;\ \overline U],
}
where $\underline X,\overline X\in\R^{Nn_x}$ and
$\underline U,\overline U\in\R^{Nn_u}$ stack the state and input bounds over the horizon, so $G$ has $m=N(n_x+n_u)$ rows. The constant terms independent of $U$ are omitted from the objective. We use the condensed QP formulation of LMPC in this work and recommend LQR pre-stabilization \cite[Sec.~IV]{jerez2011condensed} where needed to mitigate ill-conditioning for open-loop unstable systems.

\section{\fameqp{} Method}
\label{sec:method}

\fameqp{} implements the ADMM iteration~\eqref{eq:admm2} with the pre-inverted matrix $M=K^{-1}$, and concentrates
on how that iteration is executed on the GPU. At setup, the problem is scaled once by a modified Ruiz equilibration
\cite{ruiz2001scaling, stellato2020osqp}: $P\to cTPT$, $q\to cTq$, $G\to VGT$,
$(l,h)\to(Vl,Vh)$, with positive diagonal $T\in\R^{n\times n}$ and
$V\in\R^{m\times m}$ equilibrating the columns and rows of
$\bigl[\begin{smallmatrix}P&G^\top\\G&\mf 0\end{smallmatrix}\bigr]$ in ten
passes and $c>0$ normalizing the cost. The iteration runs on the scaled
problem, whose solution $(z',s',\lambda')$ maps back as $z=Tz'$, $s=V^{-1}s'$,
$\lambda=c^{-1}V\lambda'$; the test~\eqref{eq:stop} is evaluated on the
unscaled residuals, and later problem data are scaled with the same $T,V,c$. The penalty matrix is parameterized by a single scalar $\rho>0$ as $\bs\rho=\rho\diag(b)$. For each $i\in[1:m]$, let $b_i$, $l_i$, and $h_i$ denote the $i$th entries of $b$, $l$, and $h$, respectively. We set $b_i=10^3$ if $l_i=h_i$ (an equality constraint) and $b_i=1$ otherwise. Algorithm~\ref{alg:fame} summarizes the setup and solve phases.

\subsection{Constraint Structures}
\label{sec:structures}

\fameqp{} exploits constraint rows that act on individual decision variables
and distinguishes three structures of $G\in\R^{m\times n}$:
\emph{box}, $G=D:=\diag(d)$ with $d\in\R^n$, $d_i\neq0$ for all $i\in[1:n]$,
so $m=n$; \emph{split}, $G=[D;C]$ after a row permutation, with
$D\in\R^{n\times n}$ nonsingular diagonal and $C\in\R^{m_c\times n}$ the
remaining rows, $m_c>0$, so $m=n+m_c$; and \emph{general}, with no complete
diagonal block, where all rows are handled as dense, $C=G$ and $m_c=m$.
Under standard condensing, LMPC with input bounds
alone has the box structure and adding state bounds yields the split structure. LQR pre-stabilization \cite{jerez2011condensed} couples the input constraints and produces the general structure.
In split form, for each $i\in[n+1:m]$ (corresponding to the rows of $C$ only), we set $b_i=10^3\gamma$ if $l_i=h_i$ (equality constraints) and $b_i=\gamma$ otherwise, with $\gamma>0$.

\subsection{Two Operator Forms}
\label{sec:operators}

The first two updates in~\eqref{eq:admm2} couple different entries through
matrix--vector products; the remaining updates are elementwise because
$\bs\rho$ is diagonal and $\mc C$ is a box. Omitting the iteration index
$j$, we describe the operators for each constraint structure.

\noindent \textbf{\emph{General structure.}}
With $C=G$ and $m_c=m$, define
$r=[\sigma z-q;\;\bs\rho s-\lambda]\in\R^{n+m}$.
Then~\eqref{eq:linsys} and~\eqref{eq:linsys2} admit the \emph{stacked form}
\alnn{
[\tilde z;\tilde s]=Wr,\quad
W=[M,\;MG^\top;\;GM,\;GMG^\top],
\label{eq:W}
}
which computes both updates in one matrix--vector product.
The \emph{factored form} stores $F=[M,\;MG^\top]$ and $G$, and computes
$\tilde z=Fr$ followed by $\tilde s=G\tilde z$.

\noindent \textbf{\emph{Split structure.}}
For $G=[D;C]$, partition $\bs\rho$, $s$, and $\lambda$ conformally.
Since $D=\diag(d)$, its contribution can be absorbed elementwise into
$r=[\sigma z-q+d\circ(\bs\rho_d s_d-\lambda_d);
\;\bs\rho_c s_c-\lambda_c]\in\R^{n+m_c}$,
where $\circ$ denotes elementwise multiplication.
The stacked form~\eqref{eq:W} then applies with $G$ replaced by $C$
and $\tilde s$ by $\tilde s_c$; the factored form uses
$F=[M,\;MC^\top]$ followed by $\tilde s_c=C\tilde z$.
Both recover $\tilde s_d=d\circ\tilde z$, so $D$ need not be stored
in either operator.

\noindent\textbf{\emph{Box structure.}}
With $G=D$ and $m_c=0$, define
$r=\sigma z-q+d\circ(\bs\rho s-\lambda)$.
Both forms reduce to $\tilde z=Mr$ and $\tilde s=d\circ\tilde z$,
using one fused kernel per iteration. Only the dense matrix $M$ and
the diagonal vector $d$ are stored, requiring $n^2+n$ entries in total.
Compared with treating $D$ as dense, this reduces storage from $4n^2$
entries for the stacked form or $3n^2$ for the factored form.
After exploiting the diagonal structure, the two forms are identical.

For the split and general structures, the stacked operator contains
$(n+m_c)^2$ entries, whereas $F$ and $C$ together contain $n(n+2m_c)$.
Thus, the factored form reads
\begin{equation}
  (n+m_c)^2-n(n+2m_c)=m_c^2
  \label{eq:mc2}
\end{equation}
fewer operator entries per iteration, at the cost of two sequential
kernels instead of one fused kernel.

\subsection{The Penalty Bank: Adaptation and Probing}
\label{sec:probe}

\fameqp{} precomputes the operators for a bank of five penalties,
spaced by factors of $10$ and initially centered at $\rho_0=0.1$.
Adapting $\rho$ during a solve therefore selects a stored operator
rather than recomputing an inverse. After each convergence check,
OSQP's residual-balancing rule proposes
$\rho_{\rm new}
=\rho\sqrt{r_{\rm prim}\eta_{\rm d}/(r_{\rm dual}\eta_{\rm p})}$,
using the residual norms and scales in~\eqref{eq:stop}.
If the proposal differs from the current $\rho$ by more than a
factor of $5$, it is snapped to the nearest bank entry.
The multiplier $\lambda$ is retained across switches, since its
optimal value is independent of $\rho$.
The penalty grid remains fixed during each solve.

The same bank is also used to select the starting penalty.
At setup, each entry initializes a trial of at most $400$ iterations
from zero, with adaptation enabled. Each trial records the time and
iteration count needed to satisfy $1/100$, $1/10$, and $1$ times
the requested tolerances. Trials are compared at the tightest
tolerance reached by any trial; the fastest determines the starting
$\rho$, and its iteration count provides the estimate $j_{\rm est}$ used in
Sec.~\ref{sec:select}. Ties favor the first tested entry.
If an endpoint wins, the bank is re-centered there and probed again,
with at most two rebuilds. This search adds setup cost but no
per-solve probing.

For split constraints, input bounds are often active while state
bounds are mostly inactive, so a common penalty can overweight
the dense rows. We therefore probe
$\gamma\in\{10^{-3},\ldots,10\}$ at the selected $\rho$, then
re-probe $\rho$ with the selected $\gamma$.
The resulting $\gamma$ is fixed in every bank entry; in-solve
adaptation changes only the scalar $\rho$.

\subsection{Execution Modes}
\label{sec:modes}

Both modes perform the complete ADMM iteration~\eqref{eq:admm2}
on the GPU, fusing the elementwise updates with the operator products.

\noindent\textbf{\emph{Single-kernel iteration.}}
With the stacked operator, one fused kernel computes $Wr$ and
completes the iteration. Each thread block evaluates its assigned
output rows, then immediately applies the corresponding updates:
primal rows relax $\tilde z$ to update $z$; constraint rows form
$\zeta$, project it onto $[l,h]$ to update $s$, and update $\lambda$.
For split constraints, the primal rows also compute
$\tilde s_d=d\circ\tilde z$ and complete the diagonal constraint
updates. Each block then constructs its portion of the next $r$.
Two buffers alternate as input and output, preventing writes from
overwriting entries that other blocks are still reading.
For box constraints, the same kernel uses $M$ and handles all
constraint updates elementwise.

\noindent\textbf{\emph{Two-kernel iteration.}}
With the factored operator, the first kernel computes $\tilde z=Fr$,
updates $z$, and completes any diagonal constraint updates.
It stores $\tilde z$ for the second kernel, which computes
$\tilde s_c=C\tilde z$ and immediately performs the dense constraint
projection and dual update. Each kernel constructs the corresponding
portion of the next $r$. The kernel boundary ensures that all entries
of $\tilde z$ are available before evaluating $C\tilde z$.
Box constraints require no second kernel.

\noindent\textbf{\emph{CUDA-graph replay.}}
At setup, $j_{\rm chk}$ complete iterations followed by the
termination test~\eqref{eq:stop} are captured as a CUDA graph.
Each replay executes this sequence with one host (CPU) call, preserving
the dependencies between kernels and iterations.
The test computes $Pz$, $Gz$, and $G^\top\lambda$ on the device,
followed by two reduction kernels that produce the residual norms,
their scales $\eta_{\rm p},\eta_{\rm d}$, and a stop flag.
The host reads these five values once per replay to terminate,
declare a stall (no $10\%$ residual decrease in $12$ checks), or
adapt $\rho$ using the penalty bank.
After a penalty switch, $r$ is rebuilt before replay resumes.
Thus, fusion reduces kernels per iteration, while graph replay
reduces host dispatch and synchronization to once per
$j_{\rm chk}$ iterations.

\subsection{Automatic Mode Selection}
\label{sec:select}

\fameqp{} selects the operator form and tunes its kernel tile
configuration once per problem shape. The iteration-time model is
\begin{equation}
  t_{\rm iter}\approx n_{\rm ker}\tau+
  t_{\rm stream}(b_{\rm op}),
  \label{eq:cost}
\end{equation}
where $n_{\rm ker}\in\{1,2\}$, $\tau$ is the overhead per kernel node
in a replayed graph, and $b_{\rm op}$ is the operator size in bytes.
Once per device and precision, \fameqp{} measures $\tau$, the
host-read cost $t_{\rm read}$, and $t_{\rm stream}$ over feasible
operator sizes from \SIrange{0.25}{64}{\mega\byte}, stopping if
memory allocation fails. Outside the measured range,
$t_{\rm stream}$ is extrapolated linearly at the bandwidth of
the nearest endpoint. These measurements are cached.

For box constraints, the forms coincide and only the single-kernel
configuration is tuned. For split and general constraints, stacking
saves one launch but reads $\beta m_c^2$ additional bytes, where
$\beta$ is the number of bytes per value. Under a constant-bandwidth
approximation, the stacked form is predicted to be faster when
$m_c<\sqrt{\tau\,{\rm BW}/\beta}$, with ${\rm BW}$ denoting the
effective memory bandwidth.

An operator form is first excluded if its estimated penalty-bank
storage exceeds $60\%$ of currently free GPU memory; setup fails
if neither form fits. Among feasible forms, those predicted more
than $1.5$ times slower than the best are discarded.
The remaining forms and their tile configurations are timed inside
CUDA graphs using placeholder operators of the actual dimensions.
The single-kernel mode is retained unless the measured two-kernel
time is more than $10\%$ shorter. Thus, the model prunes the search,
while measured timings determine the final selection.
Using the measured $t_{\rm iter}$, the predicted time for
$j_{\rm est}$ iterations is
\begin{equation}
  T_{\rm graph}
  =\left\lceil\tfrac{j_{\rm est}}{j_{\rm chk}}\right\rceil
  \bigl(j_{\rm chk}t_{\rm iter}+t_{\rm check}+t_{\rm read}\bigr),
  \label{eq:T_graph}
\end{equation}
where $t_{\rm check}$ is the device-side termination-test time.

\subsection{Updates, Warm Starts, and Precision}
\label{sec:precision}

Between solves, new $q,l,h$ are scaled using the stored $T,V,c$,
and $r$ is rebuilt. Changes to $P$ or $G$ that
preserve the constraint structure trigger rescaling and in-place
re-inversion of either the full bank or, in \textit{lazy mode}, only the
active entry. Other entries are marked stale and recomputed when
selected by a penalty switch. The captured graph is retained,
and $\rho$ carries over between solves.
A warm start initializes $z,\lambda$ and reconstructs
$s=\proj_{\mc C}(Gz)$, optionally after shifting the previous
solution one stage along the horizon.

The bank is always inverted in \texttt{float64}, while operators
and iterates use \texttt{float32} or \texttt{float64}.
Single-precision residuals may stagnate near $10^{-6}$;
a stalled solve returns \texttt{solved\_inaccurate} if it satisfies
the termination test at $100$ times the requested tolerances.

The working precision is either fixed by the user or selected
automatically before the setup of Algorithm~\ref{alg:fame}. In automatic
mode, \fameqp{} estimates the condition number
$\kappa(K)=\lambda_{\max}(K)/\lambda_{\min}(K)$ by power iterations.
For target accuracy $\epsilon$ and $u_{32}=2^{-24}$, it selects
\texttt{float32} if $u_{32}\kappa(K)\le0.01\epsilon$ and \texttt{float64} if
$u_{32}\kappa(K)\ge\epsilon$. Otherwise, the setup is run in both
precisions, which are compared by $T_{\rm graph}$ in~\eqref{eq:T_graph},
preferring \texttt{float64} when its prediction is within $20\%$ of
\texttt{float32}.

\begin{algorithm}[htpb!]
\caption{\fameqp{}}
\label{alg:fame}
\renewcommand{\algorithmicrequire}{\textbf{Input:}}
\renewcommand{\algorithmicensure}{\textbf{Output:}}
\begin{algorithmic}[1]
\Require data $P,q,G,l,h$;\; tolerances
  $\epsilon_{\rm abs},\epsilon_{\rm rel}$;\; parameters $\sigma,\alpha$;\;
  check interval $j_{\rm chk}$;\; iteration limit $j_{\max}$;  precision
\Ensure $(z,s,\lambda)$ and status

\Statex[0] \textbf{Setup} (once)
\State Detect constraint structure; Ruiz-scale data
  \Comment{Sec.~\ref{sec:structures}}
\State Select execution mode
  \Comment{Sec.~\ref{sec:select}}
\State Build penalty bank; capture $j_{\rm chk}$ iterations and
  test~\eqref{eq:stop} as a CUDA graph
  \Comment{Sec.~\ref{sec:probe}, \ref{sec:modes}}
\State Probe $\rho$ (and $\gamma$ if split); re-center bank as needed

\Statex[0] \textbf{Solve} (each QP)
\State Scale new data; if $P$ or $G$ changed, refresh the bank
  fully or lazily \Comment{Sec.~\ref{sec:precision}}
\State Initialize $z,\lambda$ from zero or a warm start;
  $s\gets\proj_{\mc C}(Gz)$
\State Form $r$ as in Sec.~\ref{sec:operators}; $j\gets0$
\Repeat \Comment{one graph replay per pass}
  \For{$j_{\rm chk}$ iterations}
    \If{$m_c=0$} \Comment{box}
      \State $\tilde z\gets Mr$
    \ElsIf{stacked form} \Comment{single-kernel}
      \State $[\tilde z;\tilde s_c]\gets Wr$
    \Else \Comment{factored form, two-kernel}
      \State $\tilde z\gets Fr$;\; $\tilde s_c\gets C\tilde z$
    \EndIf
    \State $\tilde s\gets[d\circ\tilde z;\,\tilde s_c]$
      \Comment{blocks present}
    \State $z\gets\alpha\tilde z+(1-\alpha)z$;\;
      $\zeta\gets\alpha\tilde s+(1-\alpha)s+\bs\rho^{-1}\lambda$
    \State $s\gets\proj_{\mc C}(\zeta)$;\;
      $\lambda\gets\bs\rho(\zeta-s)$;\; form next $r$
  \EndFor
  \State Evaluate~\eqref{eq:stop} on the GPU; $j\gets j+j_{\rm chk}$
  \State Host reads the norms, scales, and stop flag
  \State \algorithmicif\ continuing \algorithmicthen\ adapt $\rho$
    as in Sec.~\ref{sec:probe}
  \State \algorithmicif\ $\rho$ switched \algorithmicthen\ refresh
    a stale entry; rebuild~$r$
\Until{converged, stalled, or $j\ge j_{\max}$}
\State \Return unscaled $(z,s,\lambda)$ and status
\end{algorithmic}
\end{algorithm}
\section{Experimental Results}
\label{sec:experiments}

We evaluate \fameqp{} through numerical benchmarks and closed-loop robotic control experiments in high-fidelity MuJoCo simulations.
The CPU QP baselines are OSQP \cite{stellato2020osqp},
PIQP \cite{schwan2023piqp}, ProxQP \cite{bambade2022proxqp},
Clarabel \cite{goulart2026clarabel}, and DAQP \cite{arnstrom2022daqp}.
The GPU baselines include the QP solvers cuOSQP
\cite{schubiger2020cuosqp}, ReLU-QP \cite{bishop2024reluqp},
qpax \cite{tracy2024qpax}, and CuPIQP \cite{song2026cupiqp},
and the MPC solvers CUDA MPC \cite{akbari2026cudampc} and
$\pi$MPC \cite{wu2026pimpc}.
We use the original implementations, except for CUDA MPC,
which we reimplement because its source code was unavailable.
All experiments run on a system equipped with an NVIDIA RTX 3080
Laptop GPU (48 SMs, \SI{16}{\giga\byte} VRAM) and an Intel
Core i7-11850H processor with \SI{32}{\giga\byte} RAM. Unless stated otherwise, \fameqp{} runs in \texttt{float64} with
$\sigma=10^{-6}$, $\alpha=1.6$,
$\epsilon_{\rm abs}=\epsilon_{\rm rel}=\epsilon$ for a stated tolerance
$\epsilon$, $j_{\rm chk}=10$, and $j_{\max}=2\times10^{4}$
($10^{5}$ in the closed-loop random-LMPC and Unitree G1 experiments), with penalty
adaptation and probing enabled; lazy updates are used where the QP matrices
change at every step (KQ-LMPC).

\subsection{Numerical Experiments: Random QP and LMPC}

\noindent\textbf{Dense random QPs.} We use the dense random QPs of~\cite{bishop2024reluqp}, with $n/4$ dense equality rows and $n/4$ dense one-sided inequality rows (approximately $30\%$ of the inequalities active), using five random seeds per problem size $n$. To match solution accuracy across solvers, we sweep each solver's tolerance over $\{10^{-3},\ldots,10^{-8}\}$ and select the loosest value yielding an $\ell_\infty$ solution error of at most $10^{-3}$ on all five instances, relative to Clarabel solutions computed at tolerance $10^{-10}$. We report the mean across instances of their median solve times. \fameqp{} is set up, including penalty probing, on a held-out instance of the same size. Setup time is excluded for all solvers; only solve time is reported. \Cref{fig:random_qp} shows the median across five instances of
their per-instance median cold-solve times, with min--max bands.
Using the mean of these per-instance medians for numerical comparisons,
\fameqp{} in \texttt{float64} outperforms all GPU baselines at every
tested size and all CPU baselines for $n\ge220$.
Over $n=220$--$1400$, its speedups are $1.2$--$25.9\times$ over
DAQP and $1.5$--$4.3\times$ over ReLU-QP, the best CPU and
competing GPU baselines, respectively.

Since cuOSQP's native implementation uses \texttt{float32},
we also evaluate \fameqp{} in this precision under the same
accuracy criterion. Single precision further reduces its solve
time at most sizes, although \texttt{float64} remains faster
at $n=600$ and $800$. At $n=1400$, the mean solve time is
$1.95$ ms in \texttt{float32}, compared with $5.01$ ms in
\texttt{float64} and $326$ ms for cuOSQP.

\noindent \textbf{GPU LMPC on random systems.}
Following~\cite{bishop2024reluqp}, we consider random unstable linear
systems with $n_u\in\{10,20,30,40,50\}$, $n_x=3n_u$, spectral radius
$1.05$, and horizon $N=40$, with $Q=I$, $R=0.1I$, an LQR terminal cost,
and input bounds $|u|\le1$. Initial states are chosen so that input bounds
are active on at least $30\%$ of the steps under exact MPC. We compare
\fameqp{} with ReLU-QP and CuPIQP, the two fastest GPU QP baselines in the
random-QP benchmark, and with the GPU MPC methods CUDA MPC and $\pi$MPC.
The QP solvers receive the same LQR pre-stabilized condensed QP
($n=40n_u$); the MPC methods receive the original OCP.

Each method runs $50$ closed-loop steps in \texttt{float64} after one
untimed solve at $x_0$. Where supported, step~0 reuses that solution and
later steps shift the previous solution one stage; CuPIQP accepts no
starting point and uses its native IP initialization. The QP solvers use
tolerance $10^{-3}$ in their native stopping tests: absolute and relative
for \fameqp{} and CuPIQP, and absolute only for ReLU-QP, whose thresholds
scale with $\sqrt{m}$ and $\sqrt{n}$. $\pi$MPC uses its native fixed-point
test with tolerance $10^{-6}$ at every iteration, and CUDA MPC, which has
no convergence test, its fixed budget of $500$ iterations. Timings exclude
setup and the initialization solve but include data updates, warm-start
shifts, and transfer of the first input to the host.

\Cref{tab:random_lmpc_warm_init} shows that \fameqp{} achieves the lowest
mean step time at every size, from $0.658$ to $4.23$ ms. It is
$3.2$--$7.3\times$ faster than ReLU-QP, the closest QP baseline, and
$3.2$--$13.8\times$ faster than $\pi$MPC, the fastest MPC baseline, with
smaller timing standard deviations than both at every size. Since the
stopping tests are not equivalent, we also compare the accuracy achieved:
closed-loop costs of all methods agree within $0.06\%$, and the largest
bound violation of the returned input is at most $2.0\times10^{-3}$ for
\fameqp{}, against $1.4$--$4.3\times10^{-2}$ for ReLU-QP, about $10^{-4}$
for $\pi$MPC, and below $10^{-6}$ for CuPIQP and CUDA MPC.

\begin{table}[t]
\centering
\caption{Online MPC step times [ms], mean $\pm$ standard deviation
over $50$ steps. Setup and initialization are excluded;
fastest means are bold.}
\label{tab:random_lmpc_warm_init}
\scriptsize
\setlength{\tabcolsep}{2pt}
\resizebox{\columnwidth}{!}{%
\begin{tabular}{@{}lccccc@{}}
\toprule
Solver / $n$ & 400 & 800 & 1200 & 1600 & 2000 \\
\midrule
\fameqp{}
& $\mathbf{0.658\pm0.066}$
& $\mathbf{1.19\pm0.04}$
& $\mathbf{2.05\pm0.45}$
& $\mathbf{2.92\pm0.09}$
& $\mathbf{4.23\pm0.21}$ \\
ReLU-QP
& $2.10\pm0.37$ & $6.06\pm1.02$ & $11.6\pm1.9$
& $20.0\pm3.9$ & $30.9\pm6.1$ \\
CuPIQP
& $23.7\pm7.0$ & $58.9\pm11.3$ & $178\pm42$
& $248\pm63$ & $449\pm120$ \\
\midrule
CUDA MPC
& $22.7\pm0.1$ & $59.9\pm0.5$ & $82.8\pm0.8$
& $103\pm0$ & $138\pm3$ \\
$\pi$MPC
& $9.06\pm12.39$ & $8.90\pm12.88$ & $6.78\pm5.67$
& $15.4\pm23.9$ & $13.4\pm18.1$ \\
\bottomrule
\end{tabular}%
}
\end{table}

\subsection{Closed-Loop LMPC on MuJoCo Robotic Systems.} 

\noindent \textbf{Unitree G1 humanoid push recovery.}
We evaluate balance recovery of the Unitree G1 in MuJoCo
(29 position-controlled joints, 70 states) at a
\SI{2}{\milli\second} control period.
The prediction model is the discrete linearization of the full
MuJoCo step about a bent-knee stance. The cost penalizes CoM
displacement and joint deviations, with an LQR terminal cost
and pre-stabilizing gain. Joint-target offsets are bounded by
rated torque divided by actuator gain
(\SIrange{0.01}{0.28}{\radian}).
The resulting condensed QP has $n=m=29N$ and general constraint
structure. We compare against ReLU-QP, the fastest competing
GPU QP solver in the random-QP benchmark. Both use
\texttt{float64} and tolerance $10^{-4}$, with convergence checked
every $10$ and $25$ iterations, respectively.

\begin{figure}[htpb!]
\centering
\includegraphics[width=0.8\columnwidth]{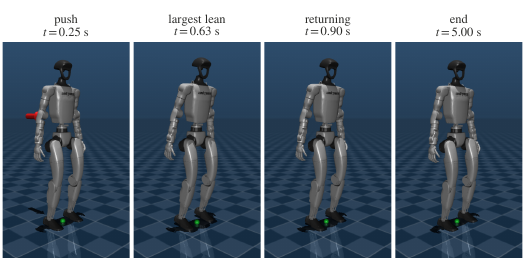}
\caption{Unitree G1 push recovery in MuJoCo with \fameqp{}: torso push,
maximum lean, return, and final stance.}
\label{fig:g1_balance_shifted}
\vspace{-0.7em}
\end{figure}

An \SI{80} N torso push lasting \SI{0.1}{\second}
tests recovery over \SI{5}{\second}
(\Cref{fig:g1_balance_shifted}).
\fameqp{} reduces mean QP time by $2.3$--$3.1\times$,
remaining below the \SI{2}{\milli\second} control period
at every tested horizon, whereas ReLU-QP exceeds it for $N\ge15$.
Thus, \fameqp{} doubles the largest tested horizon within the
mean-time budget, from $N=10$ to $N=20$.
Both methods achieve comparable balance recovery, with CoM RMS
deviations of \SIrange{18.7}{20.5}{\milli\meter}
(\Cref{tab:g1_balance_shifted}).

\begin{table}[htpb!]
\centering
\caption{Unitree G1 push recovery. QP times are mean $\pm$ std,
excluding the first step.}
\label{tab:g1_balance_shifted}
\scriptsize
\setlength{\tabcolsep}{2.5pt}
\renewcommand{\arraystretch}{0.9}
\begin{tabular}{@{}rcccc@{}}
\toprule
$N$ ($n$) & \multicolumn{2}{c}{QP time [ms]}
& \multicolumn{2}{c}{CoM RMS [mm]} \\
\cmidrule(lr){2-3}\cmidrule(lr){4-5}
& \fameqp{} & ReLU-QP & \fameqp{} & ReLU-QP \\
\midrule
10 (290) & $\mathbf{0.650\pm0.091}$ & $1.47\pm0.05$ & 18.7 & 18.7 \\
15 (435) & $\mathbf{0.763\pm0.116}$ & $2.26\pm0.12$ & 19.1 & 19.2 \\
20 (580) & $\mathbf{1.04\pm0.20}$   & $3.25\pm0.13$ & 20.5 & 19.5 \\
\bottomrule
\end{tabular}
\end{table}

\noindent\textbf{Quadrotor trajectory tracking in MuJoCo.}
We evaluate \fameqp{} with changing prediction matrices using
KQ-LMPC~\cite{rajkumar2025real}, which constructs time-varying
LMPC from analytically derived Koopman-lifted dynamics.
The Skydio X2 tracks the supplied lemniscate trajectory for
\SI{20}{\second}, starting on the reference, at \SI{50}{\hertz}.
With four inputs and $N=70$, the condensed QP has $n=280$
variables, and a \SI{1.4}{\second}
prediction horizon. Both the Hessian and the constraint matrix
change at every step.

We compare against CuPIQP, the second-fastest GPU QP baseline
in the random-QP benchmark. ReLU-QP is excluded because its
native interface does not support matrix updates between solves.
Both solvers use \texttt{float64} and tolerance $10^{-4}$,
with setup and one initial solve performed before timing.
\fameqp{} uses shifted warm starts; CuPIQP computes its own
IP starting point at each solve.

\begin{figure}[htpb!]
\centering
\includegraphics[width=0.7\columnwidth]{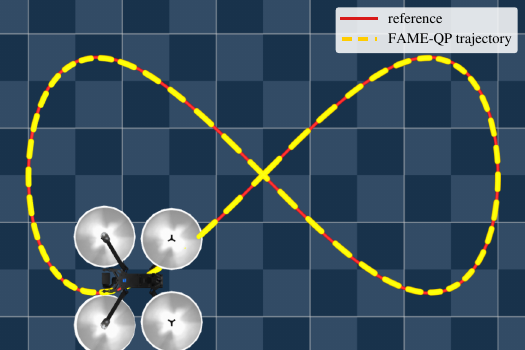}
\caption{Skydio X2 trajectory tracking in MuJoCo with \fameqp{}.}
\label{fig:kq_fame}
\end{figure}

Including matrix updates, warm-start shifts, and host transfer,
\fameqp{} achieves a mean QP time of $12.5\pm3.2$ ms
(mean $\pm$ standard deviation), compared with $59.2\pm4.8$ ms
for CuPIQP, a $4.7\times$ speedup. Its 99th-percentile latency
is \SI{17.6}{\milli\second}, below the \SI{20}{\milli\second}
control period, versus \SI{72.1}{\milli\second} for CuPIQP.
Despite the changing QP matrices, \fameqp{} maintains accurate
tracking, with position RMSE of \SI{2.7}{\milli\meter}
versus \SI{3.4}{\milli\meter} for CuPIQP; both have a maximum
tracking error of \SI{8.5}{\milli\meter}. \Cref{fig:kq_fame} shows the Skydio X2 quadrotor tracking the lemniscate trajecotry with \fameqp{}.

\subsection{Ablation Study}
\label{sec:ablations}

We isolate GPU execution, constraint structure, operator selection,
penalty probing, and matrix updates. Fixed-budget experiments time $200$
iterations at a fixed penalty without termination checks; complete solves
include convergence checks and transfer of the solution to the host.
Within each comparison, the problem data and all other settings are
identical.

\noindent\textbf{Fusion and graph replay.}
We compare separate operations and fused kernels, each with ordinary
launches or CUDA graph replay, on the same stacked operator with
$m_c=n/4$. Table~\ref{tab:ablations} shows that the two are complementary:
at $n=128$, replay reduces the unfused iteration time from $140$ to
$24.6\,\mu$s and fusion further to $4.37\,\mu$s, replacing $14$ kernels
per iteration with one. The gain shrinks as the operator grows and the
matrix--vector product dominates. All variants produce the same iterates
to within $3\times10^{-10}$.

\begin{table}[t]
\centering
\caption{Execution and penalty ablations. Execution:
median device time per iteration over $100$ repetitions.
Probing: mean iteration counts and mean per-instance median
wall times over five test instances, with $20$ cold solves
per instance. Setup is excluded.}
\label{tab:ablations}
\scriptsize
\setlength{\tabcolsep}{3pt}
\begin{tabular}{@{}rrrrr@{}}
\toprule
\multicolumn{5}{c}{Execution time per iteration [$\mu$s]}\\
$n$ & Unfused & Unfused & Fused & Fused\\
    & eager & graph & eager & graph\\
\midrule
128  & 140 & 24.6 & 21.1 & 4.37\\
512  & 146 & 33.4 & 21.8 & 17.4\\
1028 & 145 & 72.9 & 54.4 & 53.4\\
\midrule
\multicolumn{5}{c}{Starting-penalty probing, common bank, $\epsilon=10^{-5}$}\\
& \multicolumn{2}{c}{Iterations}
& \multicolumn{2}{c}{Solve time [ms]}\\
$n$ & Default & Probed & Default & Probed\\
\midrule
250  & 74  & 46 & 1.10 & 0.79\\
600  & 98  & 48 & 3.78 & 1.88\\
1000 & 128 & 40 & 10.6 & 3.21\\
\bottomrule
\end{tabular}
\end{table}

\noindent\textbf{Constraint structure.}
For QPs with $G=[I;C]$, $m_c\in\{0,n/4,n\}$, and both operator forms, we
compare treating every row as dense against extracting the diagonal rows.
Extraction reduces operator storage by $1.67$--$4.00\times$ and iteration
time by $1.07$--$3.03\times$, with unchanged iteration counts and
solutions (within $10^{-12}$). At $n=512$ and $m_c=128$, the stacked
operator shrinks from $10.1$ to $3.12$ MiB and the iteration time from
$49.8$ to $17.3\,\mu$s, whereas ReLU-QP's $(n+2m)^2$ operator needs
$24.5$ MiB and $86.3\,\mu$s under the same graph replay; overall, it is
$1.6$--$5.8\times$ slower per iteration than the extracted operator.

\noindent\textbf{Automatic operator selection.}
We sweep $m_c$ from $16$ to $1024$ for $n\in\{128,512,1028\}$ and compare
the stacked, factored, and automatically selected forms. Automatic
selection picks the faster form at $30$ of $36$ points; the six misses
fall within the $10\%$ retention margin of Sec.~\ref{sec:select}, and the
selected configuration is at most $9.5\%$ slower than the faster fixed
form, whereas always stacking or always factoring is up to $2.2\times$ or
$1.5\times$ slower. Pruning never excludes the faster form, but the model
alone misplaces the crossover by about a factor of two, hence the final
choice by measured timings.

\noindent\textbf{Penalty probing.}
To separate the starting penalty from bank re-centering, we compare
default and probed starts on the dense random QPs, with one bank spanning
both ranges. At tolerance $10^{-5}$, the probed start reduces iterations
by $1.6$--$3.2\times$ and complete-solve time by $1.4$--$3.3\times$
(Table~\ref{tab:ablations}); widening the bank alone changes iteration
counts by at most $5\%$. A separate run crosses unfused and fused
iterations with and without the full probe (start and re-centered bank).
Iteration counts do not depend on fusion, so the gains multiply: from
$n=250$ to $1000$, fusion contributes $2.00\times$ falling to
$1.26\times$ and the probe $1.40\times$ rising to $3.35\times$, for
$2.8$--$4.2\times$ combined, with the setup cost recovered after about
$36$--$122$ solves. Probed \fameqp{} and default ReLU-QP reach $10^{-3}$
accuracy in similar iteration counts ($30$--$38$ versus $30$--$32$), so
the speedup over ReLU-QP stems mainly from per-iteration cost.

For split constraints, we replay a recorded quadrotor KQ-LMPC sequence
($N=50$) whose QP matrices change at every step. Probing $\rho$ lowers
the median and tail step times relative to the default penalty; probing
$\gamma$ as well leaves the median essentially unchanged but further
shortens the tail and nearly eliminates in-solve penalty switches.

\noindent\textbf{Lazy matrix updates.}
In KQ-LMPC, $P$ and $G$ change at every step, so the pre-inverted
operators must be rebuilt before each solve. Replaying the recorded
quadrotor QPs with identical data and warm starts, we compare full
refresh, which re-inverts all five bank entries at every step, with lazy
refresh, which re-inverts only the active entry and the others when a
penalty switch selects them. Full refresh makes the update, not the
iterations, the dominant cost of a step; lazy refresh roughly halves the
update and median step times and shortens the tail, with identical
iteration counts, solutions equal to numerical precision, and unchanged
closed-loop tracking error.
\section{Conclusion}
\label{sec:conclusion}

FAME-QP accelerates dense MPC quadratic programs through constraint-dependent operators, fused GPU execution with automatic configuration selection, penalty probing, and lazy matrix updates. Numerical and simulated robotic experiments demonstrate reduced latency while maintaining solution accuracy and control performance, including with changing prediction matrices. Future work will explore sparse and horizon-structured formulations to improve scalability.

\bibliographystyle{IEEEtran}
\bibliography{bibl}

\end{document}